\documentclass[11pt,a4paper]{article}
\usepackage[hyperref]{emnlp2020}
\usepackage{times}
\usepackage{latexsym}

\usepackage{amssymb}
\usepackage{amsmath}
\usepackage{graphicx}
\usepackage{subcaption}

\usepackage{wrapfig,lipsum,booktabs}

\usepackage{microtype}

\usepackage{multirow}

\aclfinalcopy

\title{Improving Target-side Lexical Transfer \\ in Multilingual Neural Machine Translation}

\author{
    Luyu Gao, Xinyi Wang, and Graham Neubig \\
    Language Technologies Institute,
    Carnegie Mellon University \\
    \texttt{\{luyug,xinyiw1,gneubig\}@cs.cmu.edu}
}

\date{}

\begin{document}
\maketitle
\begin{abstract}
To improve the performance of Neural Machine Translation~(NMT) for low-resource languages~(LRL), one effective strategy is to leverage parallel data from a related high-resource language~(HRL). However, multilingual data has been found more beneficial for NMT models that translate from the LRL to a target language than the ones that translate into the LRLs. In this paper, we aim to improve the effectiveness of multilingual transfer for NMT models that translate \emph{into} the LRL, by designing a better decoder word embedding. Extending upon a general-purpose multilingual encoding method Soft Decoupled Encoding~\citep{SDE}, we propose DecSDE, an efficient character n-gram based embedding specifically designed for the NMT decoder. Our experiments show that DecSDE leads to consistent gains of up to 1.8 BLEU on translation from English to four different languages.\footnote{Open-source code is available at \url{https://github.com/luyug/DecSDE}}
% \footnote{We will open source the code/data upon acceptance.}
%\footnote{Code to reproduce experiments is available at .... \gn{Remember to try to get as many items on the reproducibility checklist as possible: \url{https://2020.emnlp.org/call-for-papers}}}
\end{abstract}
\section{\label{sec:introduction}Introduction}

% \gn{Title suggestion: ``Improving Target-side Lexical Transfer in Multilingual Neural Machine Translation''? I tend to prefer to avoid abbreviations in titles.}

%Neural Machine Translation~(NMT; \citet{seq2seq}) is the current state-of-the-art method for a wide range of translation tasks \gn{this first sentence is well-known in the MT community, so it's probably not necessary}. However, its success relies on the availability of large amount of parallel data \gn{This more recent work casts doubt on the conclusions of \citet{koehn-knowles-2017-six}, so we might need to be careful in our claims: \cite{sennrich-zhang-2019-revisiting}}, which is impractical to acquire for many low-resource languages~(LRL)~\citep{koehn-knowles-2017-six}. 
The performance of Neural Machine Translation~(NMT; \citet{seq2seq}) tends to degrade on low-resource languages~(LRL) due to a paucity of parallel data~\citep{koehn-knowles-2017-six,sennrich-zhang-2019-revisiting}.
% Recent work shows that carefully chosen hyper-parameters can alleviate this problem but still demonstrate significant performance degradation with smaller data sizes~\cite{sennrich-zhang-2019-revisiting}.
One effective strategy to improve translation in LRLs is through multilingual training using parallel data from related high-resource languages~(HRL)~\citep{Zoph2016TransferLF,Neubig2018RapidAO}. The assumption underlying cross-lingual transfer is that by sharing parameters between multiple languages the LRL can benefit from the extra training signal from data in other languages. One of the most popular strategies for multilingual training is to train a single NMT model that translates in many directions by simply appending a flag to each source sentence to indicate which target language to translate into~\citep{multi_nmt_share_enc,johnson-etal-2017-googles}.

Many works focus on using multilingual training to improve \emph{many-to-one} NMT models that translate \emph{from} both an HRL and an LRL to a single target language~\citep{Zoph2016TransferLF,Neubig2018RapidAO,gu-etal-2018-meta}. In this situation, sentences from the HRL-target corpus provide an extra training signal for the decoder language model, on top of cross-lingual transfer on the source side. When training an NMT model that translates \emph{into} an LRL, however, multilingual data tends to lead to smaller improvements~\citep{multi_for_low_resource,massive_wild,massive}.
% \gn{Commented this out because it's repetitive and we'll want to save space.}
% They find that positive transfer from the HRLs to the LRLs is much less for the model that translates into the LRLs than the reverse direction.  

%\gn{I think somewhere it'd probably be a good idea to mention why we focus on word embeddings rather than, for example, the decoder transformer/LSTM parameters.}

In this paper, we aim to improve the effectiveness of multilingual training for NMT models that translate \emph{into} LRLs. Prior work has found vocabulary overlap to be an important indicator of whether data from other languages will be effective in improving NMT accuracy~\citep{tcs,lin-etal-2019-choosing}. Therefore, we hypothesize that one of the main problems limiting multilingual transfer on the target side is that the LRL and the HRL may have limited vocabulary overlap, and standard methods for embedding target words via lookup tables would map corresponding vocabulary from these languages to different representations.

To overcome this problem, we design a target word embedding method for multilingual NMT that encourages similar words from the HRLs and the LRLs to have similar representations, facilitating positive transfer to the LRLs. While there are many methods to embed words from characters \cite{wang_ling_char_nmt,char_cnn,wieting-etal-2016-charagram,ataman-federico-2018-compositional}, we build our model upon Soft Decoupled Encoding~(SDE; \citet{SDE}), a recently-proposed general-purpose multilingual word embedding method that has demonstrated superior performance to other alternatives. SDE represents a word by combining a character-based representation of its spelling and a lookup-based representation of its meaning. %\gn{``semantic embedding'' is vague, and this is also leaving out the language specific transform.}. 
We propose DecSDE, an efficient adaptation of SDE to NMT decoders. DecSDE uses a low-rank transformation to assist multilingual transfer, and it precomputes the embeddings for a fixed vocabulary to speedup training and inference. 
We test our method on translation from English to 4 different low-resource languages, and DecSDE brings consistent gains of up to 1.8 BLEUs.

\section{Translating into Low-resource Languages}
Standard NMT training is performed solely on parallel corpora from a source language $S$ to a target language $T$. However, in the case that $T$ is an LRL, we can use parallel data from $S$ and a related HRL $T'$ to assist learning. 
%\gn{This whole paragraph is basically duplicating content in the intro. Could you replace it with something else? For example, a mathematical definition of the standard lookup embeddings, similar to what you do for SDE, and then a brief re-iteration of why this will fail to encourage transfer. There were also some other places in the paper where you repeat the same content multiple times. It's a really good idea to avoid duplication unless you \emph{really} think that the reader might have forgotten what you said earlier, and in that case you can say something like ``as mentioned in the introduction'' to remind the reader that you've already covered it but are reminding them again because it's an important point.}
The standard look-up embedding in NMT turns words from both the LRL and the HRL into vectors by mapping their indices in the vocabulary to the corresponding entry in the embedding matrix. This is harmful for positive transfer, because different words with similar spellings from the LRL and the HRL are mapped to independent embeddings. For example, ``Ola'' in Galician and ``Olá'' in Portuguese both mean ``hello'', but they would have separate representations through the look-up embedding.  
We give a demonstration of this embedding (mis-)alignment in \autoref{sec: embedding-analysis}. 
Since the target side data is essential for training the decoder's language model, representing lexicons from the LRL and HRL into shared space is especially important to improve positive transfer for NMT models that translate into LRLs.
\section{Soft Decoupled Encoding}
\label{sec:method}

% \gn{I hadn't checked the math carefully here yet, but there are several issues both related to consistency and correctness. (1) It is not clear if you are using row or column vectors. ``We extract a bag of $n$-grams frequency vector from $w$, denoted as $\text{BoN}(w)$, where each row corresponds to the number of times a character $n$-gram in the vocabulary appears in $w$.'' indicates that they should be column vectors (because only column vectors have multiple rows), but then you are taking vector-matrix products $\text{BoN}(w) \cdot \textbf{W}_{c}$, which can only be done with respect to row vectors. I personally prefer column vectors and matrix-vector products (which I think are more standard in math and NLP, although there are exceptions such as \citet{transformer}), but you should definitely be consistent either way. (2) You are using the $\cdot$ symbol, $\times$ symbol, and no symbol all interchangeably to represent matrix multiplication. I think ``no symbol'' is the standard way to do this in mathematics, but I think the $\times$ symbol should also be OK for clarity if you're multiplying between matrices and function outputs. $\cdot$ probably should be reserved for dot product. (3) You're using $D$ and $d$ in different places. (4) You're using ``text'' for some functions (tanh, BoN) and ``texttt'' for others (softmax), try to use just ``text''. (5) If you want more info on typesetting math, this is a good reference: \url{http://demo.clab.cs.cmu.edu/cdyer/short-guide-typesetting.pdf}}

To address the limitation of the standard word representation for target side multilingual transfer, we turn to Soft Decoupled Encoding~(SDE;\citet{SDE}), a word embedding method designed for multilingual data. SDE decomposes a word embedding into two components: a character n-gram embedding with a language-specific transformation that represents its spelling, and a semantic embedding that represents its meaning. Given a word $w$ from the target language $L_i$, SDE embeds the words in three steps.
%In this section, we introduce our proposed method of encoding output pieces. In particular, we will discuss how we build the underlying n-gram embedding, how they are used to estimate piece embeddings, and finally, how is the embeddings matrix of pieces used in an NMT model. 

%\subsection{Character N-gram Embedding}
%When using SDE in encoder, words are mapped to a finite character vocabulary \cite{SDE}. In comparison, one challenge of using SDE for decoding process is that the decoder decodes to a infinite output space. To address this problem, we first learn piece segmentation and apply SDE over the pieces. Given a parallel corpus, we run SentencePiece \cite{SentencePiece} to generate a shared piece vocabulary of size $M$. We then extract n-grams with sliding window of size from 1 to $L$ and collect the most frequent $C$ n-grams, which are set as the n-gram Vocabulary. A trainable n-gram embedding matrix $\textbf{W}_{ng} \in \mathbb{R}^{C \times d}$ is maintained, in which each row is a embedding vector that correponds to one of the n-gram. During training, in forward pass this embedding matrix is used to estimate piece embedding, while in backward pass gradients are back-propagated to update it.
%
%\subsection{Building Piece Embedding}
%To estimate the piece embedding, we aggregate its n-gram embeddings, and apply a series of transformations. These piece embeddings are stacked to build the piece embeddings matrix.

\paragraph{Character aware embeddings} \label{sec: charater aware embedding} are first used to calculate the lexical representation of $w$.
We extract a bag of $n$-grams frequency vector from $w$, denoted as $\text{BoN}(w)$, where each row corresponds to the number of times a character $n$-gram in the vocabulary appears in $w$. 
%We normalize this n-gram vector by L1 norm to get $\hat{\text{BoN}}(p) = \text{BoN}(p) / \sum_{i=1}^C \text{BoN}_i(p)$. 
The character aware embedding of the $w$ is then computed as
\begin{equation}
    c(w) = \text{tanh} \left(\textbf{W}_{c} \; \text{BoN}(w)\right),
\end{equation}
where $\text{tanh}$ is the activation function and $\textbf{W}_{c} \in \mathbb{R}^{d \times n}$ is an embedding matrix of dimension $d$ for the $n$ character n-grams in the vocabulary.

\paragraph{Language-specific transformation} is then applied to lexical embedding $c(w)$ to account for the divergence between the HRL and the LRL: 
\begin{equation}
    \label{eq: LT}
    c_i(w) = \text{tanh} \left(\textbf{W}_{L_i} \; c(w) \right),
\end{equation}
where the matrix $\textbf{W}_{L_i} \in \mathbb{R}^{d \times d}$ is a linear transformation specific to the language $L_i$.

\paragraph{Latent semantic embeddings} of $w$ are calculated using an embedding matrix $W_s \in \mathbb{R}^{d \times s}$ with $s$ entries, which is shared between the languages. We use $c_i(w)$ as the query vector to perform attention~\citep{attention} over the embeddings
\begin{equation}
    s(p) = \textbf{W}_s \; \text{softmax}\left(\textbf{W}_s^\top \; c_i(w) \right).
\end{equation}
The final embedding of $w$ is obtained by summing the lexical and semantic representations
\begin{equation}
    e_\text{SDE} (w) = c_i(w) + s(w).
\end{equation}

\section{DecSDE for NMT Decoders}
In this section, we build upon the previously described SDE, and design a new method for multilingual word representation on the \emph{target} side. 

There are two aspects to consider when incorporating character-based representations like SDE in decoders: 1) the embedding method should be efficient during both training and inference time, as it needs to be calculated over the entire vocabulary; 2) it should support popular decoder design decisions, such as weight tying~\citep{WeightTying}, which allows the decoder to share the parameters of the target embedding matrix and the decoder projection before the softmax operation. With these considerations in mind, we introduce DecSDE, a multilingual target word embedding method based on SDE for NMT decoders.
%\gn{It's a bit strange that you lay out two problems (efficiency and supporting weight tying), then you talk about something that aims just to improve performance. Try to adjust the desiderata noted above to match the argumentation in the paragraphs below. In terms of priority/order: I think maybe ``efficient inference'', ``weight tying'', ``low-rank'' would be good as the first two are really decoder-specific, and the final one is something that may not be specific to the decoder.}

%\cw{the following sentence is obvious from the math formulat. say something more intuitive about why low-rank is necessary.}Here, $\textbf{U}_{L_i}$ projects the embedding to a $\mathbb{R}^{u}$ subspace to get the language-specific component to be transformed; $\textbf{V}_{L_i}$ performs transformation and project the embedding back to $\mathbb{R}^{d}$. 
%Meanwhile, Identity in Eq.~\ref{eq: LLT} carry over component common among languages.

\paragraph{Fixed Vocabulary and Weight Tying} The standard SDE is designed to encode words directly without segmenting them into subwords~\citep{SDE}. This design choice works well for encoding words on the source side, but it can cause problems for the decoder, which requires a finite vocabulary to generate words for each time step. Therefore, we choose to segment the target sentences into subwords~\citep{SentencePiece}, and encode each subword using DecSDE. 

The use of a fixed vocabulary also allows us to perform weight tying. Specifically, we construct an embedding matrix for the decoder by precomputing the DecSDE embedding for each subword in the target vocabulary. This embedding matrix can then be used both as the encoder lookup table and as the projection matrix before the decoder softmax.  
%such as Use of fixed vocabulary allows us to precompute the DecSDE embedding for each piece in the target vocabulary and construct the embedding matrix. Therefore, it allow us to put global regularization on the embedding. In DecSDE, we employ an indirect weight tying where the precomputed DecSDE target embeddings are used as the projection matrix before the softmax operation.

\paragraph{Efficient Training and Inference} One drawback of the standard SDE is that it requires more computation than standard look-up table embeddings because the lexical embedding requires one to extract and embed all character n-grams for each word. This problem is especially important for the decoder, since it needs to embed all target words in the vocabulary for each time step to calculate the probability distribution over the vocabulary. 

To make training more efficient, we extract the character n-grams for all words in the target vocabulary, and use an optimized embedding bag layer\footnote{Implementation with \text{torch.nn.functional.embedding\_bag}} to parallelize the calculation of lexical embeddings for all words in a batch. For inference, we precompute the DecSDE embedding for all subwords, effectively making inference as fast as the regular look-up table embedding. An analysis of training and inference speed can be found in \autoref{sec:experiment-result}.
%TO alleviate the problem, for training, we leverage GPU's parallelism: we extract character n-grams for all target pieces, store them in a tensor, and in forward pass use an embedding bag layer\footnote{Implementation with \text{torch.nn.functional.embedding\_bag}} to parallellize the calculation of lexical embedding of all pieces. Since DecSDE use a fixed vocabulary, for inference, we precompute and reuse DecSDE embedding for all subwords, effectively making inference as fast as the regular lookup table embedding.
% Since DecSDE use a fixed vocabulary, 
% we can parallely compute the DecSDE embeddings for all subwords in the vocabulary to speedup its training; for inference, DecSDE embeddings remain unchanged and are pre-computed, allowing decoding speed to be as fast as the regular lookup embedding. Detailed
% implementation and speed comparison can be found in \autoref{sec:speed} 
\paragraph{Low-rank Language-Specific Transformation}
The language-specific transform in the standard SDE used on the encoder side sometimes hurts the model performance~\cite{SDE}. Our experiments confirm that this phenomenon also happens on the decoder side. We hypothesize that this is because the full-rank transformation matrix, that is $\textbf{W}_{L_i}$ in \autoref{eq: LT} might overfit the training data and project the lexical embeddings from different languages too far from each other, which could hurt multilingual transfer.    
%We believe that transformation matrices $\textbf{W}_{L_i}$  overfits the data and transform shared character embedding sub-components, thus hurting multilingual transfer.
Therefore, we introduce a novel low-rank language-specific transformation for DecSDE: We upper-bound the rank of the transformation matrix so that it is less complex, which can encourage generalization. Specifically, we replace language-specific transformation matrix $\textbf{W}_{L_i}$ in \autoref{eq: LT} with two components: an identity matrix and a low-rank factorized matrix,
\begin{equation}
    \label{eq: LLT}
    %\textbf{W}_{L_i} = \textbf{I} + \textbf{T} \text{, where } \textbf{T} = \textbf{U}_{L_i} \textbf{V}_{L_i}
    \textbf{W}_{L_i} = \textbf{I} + \textbf{U}_{L_i} \textbf{V}_{L_i}
\end{equation}
where $\textbf{U}_{L_i} \in \mathbb{R}^{d \times u}$ , $\textbf{V}_{L_i} \in \mathbb{R}^{u \times d}$ are the low-rank matrices with dimension $u < d$. Thus, the identity matrix $\textbf{I}$ passes through the lexical embedding as-is, and the low-rank matrix performs a simple transformation to account for the divergence between languages without amplifying the difference.  
\paragraph{Extension to Multiple Target Language}
Note that though in this work we focus on HRL and LRL pairs, one can easily extend the framework to multiple~($>2$) target languages. In particular, the only language dependent component of DecSDE is the matrices $W_{L_i}$, while the rest of DecSDE parameters as well as transformer encoder-decoder parameters are shared. We can add and train $W_{L_j}$ for each of additional language $L_j$.

\section{Experiments}
\label{sec:experiment-method}
%\gn{If I remember correctly, at some point you had a comparison of using different sizes of subword vocabulary with Lookup-Piece and DecSDE. I actually though that was pretty interesting, but it looks like it's been removed. I think it would be worth adding this, and there's still a lot of space to compress the paper, for example by using ``vspace\{1mm\} noindent textbf'' instead of ``paragraph'', making Figure 1 more compact (e.g. move the legend within the $\delta$MRR figure, and maybe making both figures shorter), by reducing verbosity of the text, etc. Perhaps you could re-add this?}
\subsection{Setup}
\paragraph{Datasets}
To validate our method, we use the 58-language-to-English TED corpus for experiments \citep{qi-etal-2018-pre}. We use three LRL datasets: Azerbaijani (aze), Belarusian (bel), Galician (glg) to English, and a slightly higher-resource dataset Slovak (slk). Each LRL is paired with a related HRL: Turkish (tur), Russian (rus), Portuguese (por), and Czech (ces) respectively. We translate from English to each of the four LRLs, and train together with the corresponding HRL. For simplicity, as a research setup, we do not use back-translation with mono-lingual data which is also hard to come by for languages low in resource we experiment with.

\paragraph{Implementation}
We implement our method using the fairseq~\cite{fairseq} toolkit.  We use the Transformer~\cite{transformer} NMT model with 6 encoder and decoder layers and 4 attention heads. Other details of model architecture can be found in \autoref{app:train}. 
For all experiments, we use SentencePiece~\cite{SentencePiece} with a vocabulary size of 16K.
%As baseline, we use the same set of pieces but a lookup based word embedding in the decoder and tie it with output weight matrix.
%For fair comparison, we keep the rest of the model architecture the same. In particular, a single transformer with six encoder layers and six decoder layers is shared by the low resource and high resource language pair.

\paragraph{Compared Systems}
We compare with two systems:
1)~LookUp-piece: we use SentencePiece separately on each language to get subword vocabularies. Both encoder and decoder use look-up based embeddings.
2)~LookUp-word: We concatenate the training data together and extract the most frequent 64K tokens as the shared vocabulary. Both encoder and decoder use look-up based embeddings. Both systems employ vanilla weight-tying.
% 3)~SDE: we also use Sentencepiece for the data. The encoder uses a lookup embedding and the decoder use the standard SDE embedding on word pieces \gn{This is actually really confusing. I thought the whole point of the paper was to use SDE on the decoder side. Why is this a baseline? What is different from DecSDE?}.
% \gn{Define DecSDE as well. Also, I know space is tight, but it'd be nice to explain what the comparison between all of these systems is mean to test.}
\subsection{Experiment Results}
\label{sec:experiment-result}

% \gn{Maybe combine section 5.2, 5.3, 5.4 into one? It would save sapce.}

\begin{table}[h]
\centering
\resizebox{\columnwidth}{!}{
\begin{tabular}{ l | c c c c}
\hline
\textbf{Model} & \textbf{aze} & \textbf{bel} & \textbf{glg} & \textbf{slk} \\
\hline
LookUp-word  & 0.26 & 2.65& 5.91 & 6.7 \\
LookUp-piece & 5.18 & 9.81 & 21.86 & 21.34 \\
\hline
DecSDE & \textbf{6.66} & \textbf{11.56} & \textbf{23.68} & \textbf{22.55} \\
- weight tying & 5.75 & 9.5 & 22.22 & 21.2 \\
- transform & 6.26 & 10.18 & 23.68 & 22.4 \\
- low-rank transform & 5.65 & 11.36 & 22.1 & 22.2 \\
% Piece-SDE w\textbackslash o LT & & & & \\
\hline
\end{tabular}}
\caption{Model performance and ablations. DecSDE outperforms the best baseline for all four languages.} %\gn{Combine Table 2 with this table?}}
\label{tab:perf and ablation}
\end{table}
\paragraph{Performance}We measure model performance using SacreBLEU \cite{sacrebleu} and summarize the results in \autoref{tab:perf and ablation}. 
%We record SDE model performance obtained with the the optimal n-gram size. Our experiments also reveal that including language specific transformation may within the SDE embedding may hurt performance and in table \ref{tab:perf} record the best setup. We study the effects of n-gram size and language specific transformation in section \ref{sec: ngram} and section \ref{sec: abl}. We keep a piece vocabulary size of 16k for each language and a latent size of 10k and effects of varying them are discussed in following section \ref{sec: piece vocab} and \ref{sec: latent}.\\
DecSDE consistently improves over the best baseline for all languages, outperforming LookUp-piece by up to 1.8 BLEU. Meanwhile, we see word-level baseline has inferior performance, likely due to little word-level overlap between HRL and LRL.
% Compared with the standard SDE, our method still achieves up to 1.86 improvement in BLEU, because DecSDE applies weight tying~\citep{WeightTying} for the decoder and has a more effective language specific transformation. 
%Meanwhile, all other methods have large margin over LookUp-word and we believe this is due to the fact that the problem with insufficiently learnt embedding as well as unknown words.
% \subsection{Effect of Low-rank Transformation}
% \label{sec: abl}
%In this section we remove each component of DecSDE to examine their importance to the model. The results are listed in \autoref{tab:abl}.
%\cw{emphsize that you are examining the low-rank transformation, say something about the results}
\paragraph{Ablation} We examine the effect of DecSDE components by removing each of them, as in \autoref{tab:perf and ablation}. First, we can see that removing weight tying degrades the model performance by a large margin for all four languages. Next, comparing the standard linear transformation~(- low-rank transform), and the method without the entire language-specific transform component~(- transform), we can see that using the regular transform without low-rank factorization actually degrades the model performance for three out of the four languages, indicating that a full linear transformation might hinder multilingual transfer. Using the low-rank transform achieves the best performance for all four languages. 
\paragraph{Speed} 
We measure the training time for one epoch, and the decoding time of the whole test set for aze. The results are in \autoref{tab:train-test-speed}. DecSDE incurs a reasonable training overhead, and has similar inference speed as the regular lookup embedding.

%\begin{wraptable}{r}{0.3\textwidth}
\begin{table}
\centering
% \resizebox{\columnwidth}{!}{
\begin{tabular}{ l | c c}
\hline
\textbf{Model} & \textbf{Train} & \textbf{Decode} \\
\hline
LookUp-Piece & 152 sec & 13.2 sec \\
DecSDE & 341 sec & 11.5 sec \\
\hline
\end{tabular} %}
\caption{Train/inference speed. DecSDE has similar inference speed as standard look-up embeddings.}
\label{tab:train-test-speed}
\end{table}
%\end{wraptable}

\paragraph{Effect of Vocabulary Size}
\label{sec: piece vocab}
%SentencePiece generate less splits over word tokens with a larger vocabulary threshold and more splits with a smaller threshold. This leads to difference in size and forms of decoder vocabulary. 
We compared DecSDE and LookUp-piece with different vocabulary sizes to study the impact of subword segmentation and show results in \autoref{tab: vsize}. 
DecSDE consistently outperforms LookUp-piece, but both methods tend to demonstrate decreasing accuracy as the vocabulary size gets larger. 
%We keep language specific transformation matrix full for bel, and removed for other LRLs, avoiding effect of low-rank parameter $u$.
\begin{table}[h]
\centering
\resizebox{0.49\textwidth}{!}{
\begin{tabular}{l l | c c c c}
\hline
\textbf{Method} & \textbf{\# Vocab} & \textbf{aze} & \textbf{bel} & \textbf{glg} & \textbf{slk} \\
\hline
\multirow{3}{*}{LookUp-piece} 
& 8K & 6.18 & 9.2 & 22.02 & 21.92 \\
& 16K & 5.18 & 9.81 & 21.86 & 21.34 \\
& 32K & 5.03 & 8.75 & 21.27 & 20.37 \\
\hline
\multirow{3}{*}{DecSDE}  
& 8K & 6.43 & 11.57 & 23.81 & 22.92 \\
& 16K & 6.26 & 11.36 & 23.68 & 22.4 \\
& 32K & 5.36 & 10.65 & 22.85 & 20.16 \\
\hline
\end{tabular}}
\caption{Performance with Different Vocab Size.}
\label{tab: vsize}
% \label{tab:piece}
\end{table}
%\subsection{Effect of N-gram Size}
%\label{sec: ngram}
%DecSDE builds up its character n-gram vocabulary by extracting n-grams of lengths from 1 up to $L$ from the input vocabulary. Increasing $L$ leads to an increase in number of n-grams used to estimate a single token's embedding and SDE model with a larger $L$ is more expressive. Meanwhile, introducing longer n-grams also allow the SDE to model a larger range of word-units. On the other hand, expressiveness also comes with the risk of improper training. In this section, we experiment with altering this $L$ and show how the models' behaviors change accordingly.

\paragraph{Effect of N-gram Size}
\label{sec: ngram}
DecSDE builds up its character n-gram vocabulary by extracting n-grams of lengths from 1 up to $n$ from the input vocabulary. Using a larger $n$ makes the model more expressive, but it might adds more parameters the model which could lead to overfitting. In this section, we examine the effect of different $n$ values on DecSDE. The results are listed in \autoref{tab:ngram}.

\begin{table}[h]
\centering
\begin{tabular}{ c | c c c c}
\hline
\textbf{N-gram} & \textbf{aze} & \textbf{bel} & \textbf{glg} & \textbf{slk} \\
\hline
3 & 5.24 & 11.07 & 21.45 & 21.48 \\
4 & 6.16 & 11.36 & 23.35 & 22.4 \\
5 & 6.26 & 10.86 & 23.68 & 21.62 \\
% Piece-SDE w\textbackslash o LT & & & & \\
\hline
\end{tabular}
\caption{N-gram Size}
\label{tab:ngram}
\end{table}
We observe that using upto 4-gram give a huge performance improvement, while using 5-gram leads to small improve in aze and glg but small decrease in bel, slk. This suggests using character n-grams up to size 4 is enough to provide enough discriminative power for our model.

\paragraph{Effect of Latent Size}
\label{sec: latent}
\begin{table}[h!]
\centering
\begin{tabular}{ c | c c c c}
\hline
\textbf{Latent} & \textbf{aze} & \textbf{bel} & \textbf{glg} & \textbf{slk} \\
\hline
5K & 5.73 & 11.1 & 23.87 & 22.65 \\
10K & 6.26 & 11.36 & 23.68 & 22.4 \\
20K & 5.92 & 11.06 & 22.94 & 22.25 \\
% Piece-SDE w\textbackslash o LT & & & & \\
\hline
\end{tabular}
\caption{Latent Size}
\label{tab:latent}
\end{table}
We train DecSDE with different latent embedding sizes of 5K, 10K and 20K and record BLEU in \autoref{tab:latent}. We observe small differences among them for each LRL. We do find a trend that increasing the size too high will hurt performance, indicating a latent size of around 10K is sufficient while going larger is likely to incur over-fitting problem.
\paragraph{Embedding Analysis}
\label{sec: embedding-analysis}
%\begin{figure}[h!]
%\centering
%\begin{subfigure}[b]{0.45\textwidth}
%   \includegraphics[width=1\textwidth]{plots/mrr.png}
%   %\caption{MRR Gain at Different Edit Distance}
%    \label{fig: mrr}
%\end{subfigure}
%\begin{subfigure}[b]{0.45\textwidth}
%   \includegraphics[width=1\textwidth]{plots/f1.png}
%   %\caption{F-1 Gain in Rare Word}
%    \label{fig:rare-word-gain}
%\end{subfigure}
%\caption{Benefits of DecSDE compared to LookUp-piece. \textit{Top}: Gain in embedding similarity for word pairs in HRL and LRL with similar spellings. DecSDE embed similar words from HRL and LRL into more similar space.\textit{Bottom}: Gain in word accuracy F-1 over rare words in the LRLs. DecSDE brings more gains for rare words. }
%\end{figure}

% \begin{figure}
%     \centering
%     \includegraphics[width=\columnwidth ]{plots/word_fmeas.eps}
%     \caption{F-1 Gain in Rare Word}
%     \label{fig:rare-word-gain}
% \end{figure}
% \cw{label the plot and write something about it. this is the gain in word F1 of words of DecSDE over Lookup-piece according to their frequency. DecSDE does better on rare words due to better multilingual transfer. cite compare-mt since we use it for analysis}
%\paragraph{Encoding Spelling Similarity}
%\label{sec: piece retrieval}
\begin{figure}
%\begin{center}
   \includegraphics[width=0.49\textwidth]{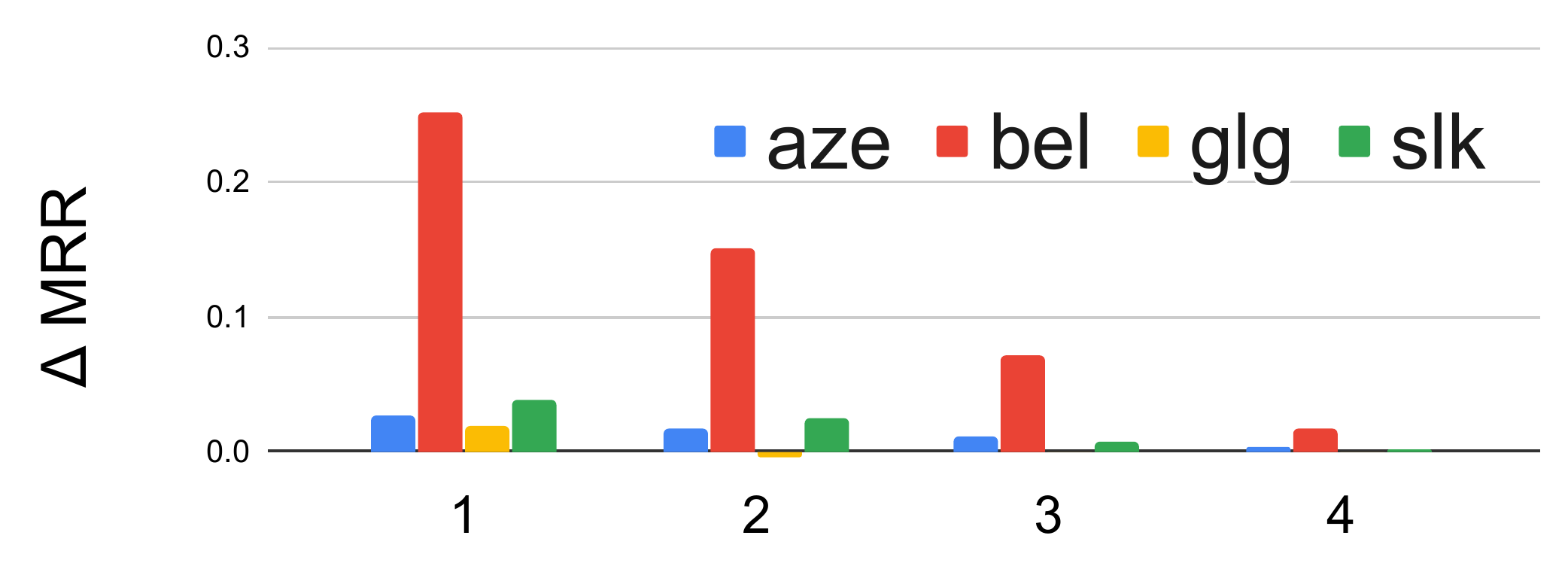}
   \includegraphics[width=0.49\textwidth]{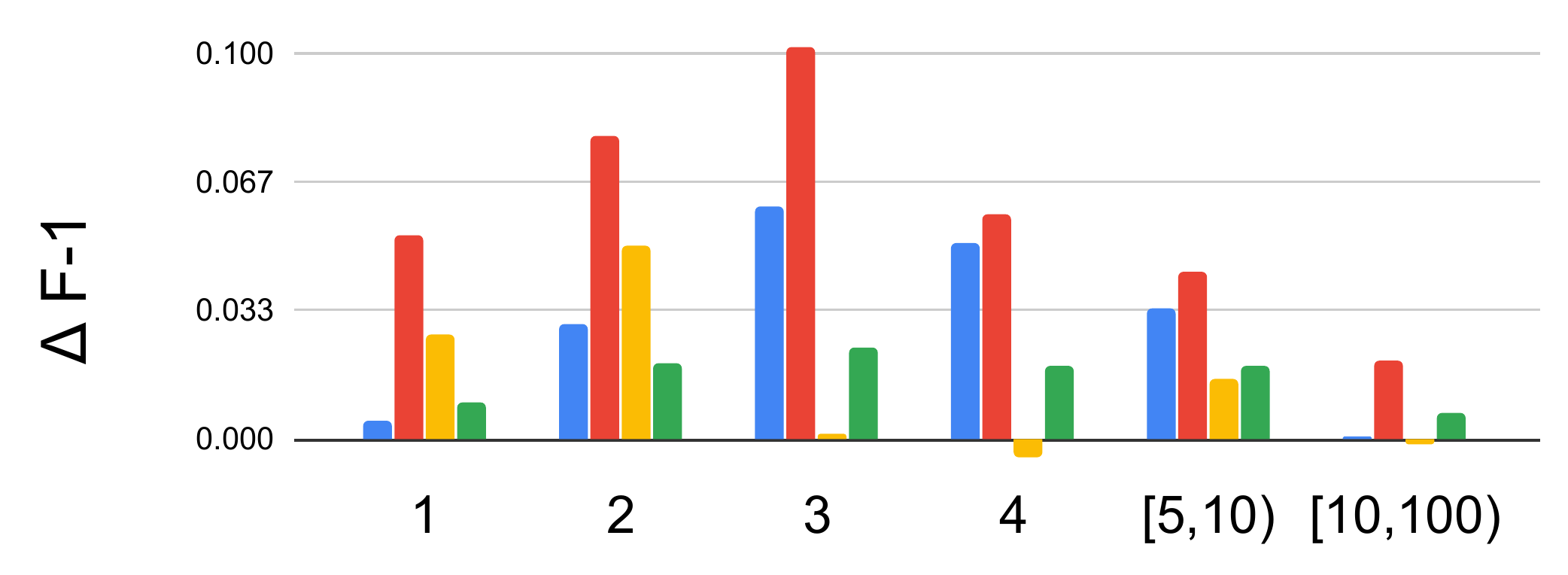}
   \vspace{-10pt}
   \captionof{figure}{\textit{Top}: Gain in embedding similarity for similarly spelled HRL, LRL word pairs. \textit{Bottom}: Gain in word accuracy F-1 over rare words in the LRLs.}
%   \gn{This figure is pixellated. Can you try to use vector graphics?}}
   \label{fig: emb-analysis}
%\end{center}
\end{figure}
One main advantage of DecSDE is its ability to capture spelling similarity
between LRL and HRL. To show this, we pick word pairs from HRL and LRL with edit distance from 1 to 4, and compare their embeddings.
% we compare the embeddings of word pairs \gn{how do you obtain these word pairs?} from the HRL and LRL with different edit distance. 
For each word pair word pair, we take the LRL word and use the cosine similarity between embeddings to retrieve words from the HRL. Retrieval success is measured by mean reciprocal rank (MRR, the higher the better). The gain of DecSDE over LookUp-piece with respect to edit distance is plotted in the top of \autoref{fig: emb-analysis}, which shows that DecSDE embed similar spelling words closer in the embedding space.

%\paragraph{Rare Word Generation}
Next, we examine performance of DecSDE for rare words in the LRLs. We calculate word F-1 of rare words for DecSDE and LookUp-piece using compare-mt~\citep{compare-mt}, and plot word frequency vs.~gain in word F-1 of in the bottom of \autoref{fig: emb-analysis}. DecSDE brings more significant gains for less frequent words, likely because it encodes similar words in HRL and LRL to closer space, thus assisting positive transfer.

% \begin{figure}[h!]
% \centering
% \includegraphics[width=\columnwidth]{plots/mrr_vs_edit.png}
% \caption{MRR Difference at Different Edit Distance}
% \label{fig: mrr}
% \end{figure}
\section{Implications and Future Work}
In this paper, we have demonstrated that DecSDE, a multilingual character-sensitive embedding method, improves translation accuracy into low resource languages.
This implies, on a higher level, that looking into the character-level structure of the target-side vocabulary when creating word or subword embeddings is a promising way to improve cross-lingual transfer.
While ablations have shown that the proposed design decisions (such as Low-rank Language-specific transformation, weight tying, etc.) are reasonable ones, this is just a first step in this direction. Future work could examine even more effective methods for target-side lexical sharing in MT or other language generation tasks.
% In this paper, we propose DecSDE, an effecient multilingual embedding method for translation into low resource language. Our contribution is 3-fold: 1) we show that using character n-gram based word embedding in decoder improves multilingual translation quality, 2) we propose a simple low rank matrix factorization technique to further assist multilingual training, and 3) we propose several implementation improvements to make it efficient for NMT decoders. Our experiments demonstrate consistent improvement over the standard look-up embedding across multiple low resource languages.

\section*{Acknowledgments}
This work was supported in part by an Apple PhD Fellowship to XW.
Any opinions, findings, and conclusions in this paper are the authors' and do not necessarily reflect those of the sponsors.

% \newpage
\bibliography{anthology,emnlp2020}

\begin{thebibliography}{26}
\expandafter\ifx\csname natexlab\endcsname\relax\def\natexlab#1{#1}\fi

\bibitem[{Aharoni et~al.(2019)Aharoni, Johnson, and Firat}]{massive}
Roee Aharoni, Melvin Johnson, and Orhan Firat. 2019.
\newblock Massively multilingual neural machine translation.
\newblock In \emph{NAACL}.

\bibitem[{Arivazhagan et~al.(2019)Arivazhagan, Bapna, Firat, Lepikhin, Johnson,
  Krikun, Chen, Cao, Foster, Cherry, Macherey, Chen, and Wu}]{massive_wild}
Naveen Arivazhagan, Ankur Bapna, Orhan Firat, Dmitry Lepikhin, Melvin Johnson,
  Maxim Krikun, Mia~Xu Chen, Yuan Cao, George Foster, Colin Cherry, Wolfgang
  Macherey, Zhifeng Chen, and Yonghui Wu. 2019.
\newblock \href {http://arxiv.org/abs/1907.05019} {Massively multilingual
  neural machine translation in the wild: Findings and challenges}.
\newblock In \emph{arxiv}.

\bibitem[{Ataman and Federico(2018)}]{ataman-federico-2018-compositional}
Duygu Ataman and Marcello Federico. 2018.
\newblock \href {https://doi.org/10.18653/v1/P18-2049} {Compositional
  representation of morphologically-rich input for neural machine translation}.
\newblock In \emph{Proceedings of the 56th Annual Meeting of the Association
  for Computational Linguistics (Volume 2: Short Papers)}, pages 305--311,
  Melbourne, Australia. Association for Computational Linguistics.

\bibitem[{Gu et~al.(2018)Gu, Wang, Chen, Li, and Cho}]{gu-etal-2018-meta}
Jiatao Gu, Yong Wang, Yun Chen, Victor O.~K. Li, and Kyunghyun Cho. 2018.
\newblock \href {https://doi.org/10.18653/v1/D18-1398} {Meta-learning for
  low-resource neural machine translation}.
\newblock In \emph{Proceedings of the 2018 Conference on Empirical Methods in
  Natural Language Processing}, pages 3622--3631, Brussels, Belgium.
  Association for Computational Linguistics.

\bibitem[{Ha et~al.(2016)Ha, Niehues, and Waibel}]{multi_nmt_share_enc}
Thanh{-}Le Ha, Jan Niehues, and Alexander~H. Waibel. 2016.
\newblock Toward multilingual neural machine translation with universal encoder
  and decoder.
\newblock \emph{Arxiv}.

\bibitem[{Johnson et~al.(2017)Johnson, Schuster, Le, Krikun, Wu, Chen, Thorat,
  Vi{\'e}gas, Wattenberg, Corrado, Hughes, and
  Dean}]{johnson-etal-2017-googles}
Melvin Johnson, Mike Schuster, Quoc~V. Le, Maxim Krikun, Yonghui Wu, Zhifeng
  Chen, Nikhil Thorat, Fernanda Vi{\'e}gas, Martin Wattenberg, Greg Corrado,
  Macduff Hughes, and Jeffrey Dean. 2017.
\newblock \href {https://doi.org/10.1162/tacl_a_00065} {{G}oogle{'}s
  multilingual neural machine translation system: Enabling zero-shot
  translation}.
\newblock \emph{Transactions of the Association for Computational Linguistics},
  5:339--351.

\bibitem[{Kim et~al.(2016)Kim, Jernite, Sontag, and Rush}]{char_cnn}
Yoon Kim, Yacine Jernite, David Sontag, and Alexander~M. Rush. 2016.
\newblock Character-aware neural language models.
\newblock \emph{AAAI}.

\bibitem[{Koehn and Knowles(2017)}]{koehn-knowles-2017-six}
Philipp Koehn and Rebecca Knowles. 2017.
\newblock \href {https://doi.org/10.18653/v1/W17-3204} {Six challenges for
  neural machine translation}.
\newblock In \emph{Proceedings of the First Workshop on Neural Machine
  Translation}, pages 28--39, Vancouver. Association for Computational
  Linguistics.

\bibitem[{Kudo and Richardson(2018)}]{SentencePiece}
Taku Kudo and John Richardson. 2018.
\newblock Sentencepiece: A simple and language independent subword tokenizer
  and detokenizer for neural text processing.
\newblock \emph{ACL}, abs/1808.06226.

\bibitem[{Lakew et~al.(2019)Lakew, Federico, Negri, and
  Turchi}]{multi_for_low_resource}
Surafel~M. Lakew, Marcello Federico, Matteo Negri, and Marco Turchi. 2019.
\newblock Multilingual neural machine translation for low-resource languages.
\newblock \emph{IJCoL}.

\bibitem[{Lin et~al.(2019)Lin, Chen, Lee, Li, Zhang, Xia, Rijhwani, He, Zhang,
  Ma, Anastasopoulos, Littell, and Neubig}]{lin-etal-2019-choosing}
Yu-Hsiang Lin, Chian-Yu Chen, Jean Lee, Zirui Li, Yuyan Zhang, Mengzhou Xia,
  Shruti Rijhwani, Junxian He, Zhisong Zhang, Xuezhe Ma, Antonios
  Anastasopoulos, Patrick Littell, and Graham Neubig. 2019.
\newblock \href {https://doi.org/10.18653/v1/P19-1301} {Choosing transfer
  languages for cross-lingual learning}.
\newblock In \emph{Proceedings of the 57th Annual Meeting of the Association
  for Computational Linguistics}, pages 3125--3135, Florence, Italy.
  Association for Computational Linguistics.

\bibitem[{Ling et~al.(2015)Ling, Trancoso, Dyer, and
  Black}]{wang_ling_char_nmt}
Wang Ling, Isabel Trancoso, Chris Dyer, and Alan~W. Black. 2015.
\newblock Character-based neural machine translation.
\newblock \emph{arxiv}.

\bibitem[{Luong et~al.(2015)Luong, Pham, and Manning}]{attention}
Thang Luong, Hieu Pham, and Christopher~D. Manning. 2015.
\newblock Effective approaches to attention-based neural machine translation.
\newblock In \emph{EMNLP}.

\bibitem[{Neubig et~al.(2019)Neubig, Dou, Hu, Michel, Pruthi, Wang, and
  Wieting}]{compare-mt}
Graham Neubig, Zi{-}Yi Dou, Junjie Hu, Paul Michel, Danish Pruthi, Xinyi Wang,
  and John Wieting. 2019.
\newblock \href {http://arxiv.org/abs/1903.07926} {compare-mt: {A} tool for
  holistic comparison of language generation systems}.
\newblock \emph{CoRR}, abs/1903.07926.

\bibitem[{Neubig and Hu(2018)}]{Neubig2018RapidAO}
Graham Neubig and Junjie Hu. 2018.
\newblock Rapid adaptation of neural machine translation to new languages.
\newblock In \emph{EMNLP}.

\bibitem[{Ott et~al.(2019)Ott, Edunov, Baevski, Fan, Gross, Ng, Grangier, and
  Auli}]{fairseq}
Myle Ott, Sergey Edunov, Alexei Baevski, Angela Fan, Sam Gross, Nathan Ng,
  David Grangier, and Michael Auli. 2019.
\newblock fairseq: A fast, extensible toolkit for sequence modeling.
\newblock In \emph{Proceedings of NAACL-HLT 2019: Demonstrations}.

\bibitem[{Post(2018)}]{sacrebleu}
Matt Post. 2018.
\newblock A call for clarity in reporting {BLEU} scores.
\newblock In \emph{Proceedings of the Third Conference on Machine Translation:
  Research Papers}, pages 186--191, Belgium, Brussels. Association for
  Computational Linguistics.

\bibitem[{Press and Wolf(2017)}]{WeightTying}
Ofir Press and Lior Wolf. 2017.
\newblock Using the output embedding to improve language models.
\newblock \emph{ArXiv}, abs/1608.05859.

\bibitem[{Qi et~al.(2018)Qi, Sachan, Felix, Padmanabhan, and
  Neubig}]{qi-etal-2018-pre}
Ye~Qi, Devendra Sachan, Matthieu Felix, Sarguna Padmanabhan, and Graham Neubig.
  2018.
\newblock \href {https://doi.org/10.18653/v1/N18-2084} {When and why are
  pre-trained word embeddings useful for neural machine translation?}
\newblock In \emph{Proceedings of the 2018 Conference of the North {A}merican
  Chapter of the Association for Computational Linguistics: Human Language
  Technologies, Volume 2 (Short Papers)}, pages 529--535, New Orleans,
  Louisiana. Association for Computational Linguistics.

\bibitem[{Sennrich and Zhang(2019)}]{sennrich-zhang-2019-revisiting}
Rico Sennrich and Biao Zhang. 2019.
\newblock \href {https://doi.org/10.18653/v1/P19-1021} {Revisiting low-resource
  neural machine translation: A case study}.
\newblock In \emph{Proceedings of the 57th Annual Meeting of the Association
  for Computational Linguistics}, pages 211--221, Florence, Italy. Association
  for Computational Linguistics.

\bibitem[{Sutskever et~al.(2014)Sutskever, Vinyals, and Le}]{seq2seq}
Ilya Sutskever, Oriol Vinyals, and Quoc~V. Le. 2014.
\newblock Sequence to sequence learning with neural networks.
\newblock In \emph{NIPS}.

\bibitem[{Vaswani et~al.(2017)Vaswani, Shazeer, Parmar, Uszkoreit, Jones,
  Gomez, Kaiser, and Polosukhin}]{transformer}
Ashish Vaswani, Noam Shazeer, Niki Parmar, Jakob Uszkoreit, Llion Jones,
  Aidan~N. Gomez, Lukasz Kaiser, and Illia Polosukhin. 2017.
\newblock Attention is all you need.
\newblock \emph{ArXiv}, abs/1706.03762.

\bibitem[{Wang and Neubig(2019)}]{tcs}
Xinyi Wang and Graham Neubig. 2019.
\newblock Target conditioned sampling: Optimizing data selection for
  multilingual neural machine translation.
\newblock In \emph{ACL}.

\bibitem[{Wang et~al.(2019)Wang, Pham, Arthur, and Neubig}]{SDE}
Xinyi Wang, Hieu Pham, Philip Arthur, and Graham Neubig. 2019.
\newblock Multilingual neural machine translation with soft decoupled encoding.
\newblock \emph{ICLR}, abs/1902.03499.

\bibitem[{Wieting et~al.(2016)Wieting, Bansal, Gimpel, and
  Livescu}]{wieting-etal-2016-charagram}
John Wieting, Mohit Bansal, Kevin Gimpel, and Karen Livescu. 2016.
\newblock \href {https://doi.org/10.18653/v1/D16-1157} {{C}haragram: Embedding
  words and sentences via character n-grams}.
\newblock In \emph{Proceedings of the 2016 Conference on Empirical Methods in
  Natural Language Processing}, pages 1504--1515, Austin, Texas. Association
  for Computational Linguistics.

\bibitem[{Zoph et~al.(2016)Zoph, Yuret, May, and Knight}]{Zoph2016TransferLF}
Barret Zoph, Deniz Yuret, Jonathan May, and Kevin Knight. 2016.
\newblock Transfer learning for low-resource neural machine translation.
\newblock In \emph{EMNLP}.

\end{thebibliography}
\bibliographystyle{acl_natbib}

\clearpage
\appendix
\section{Appendix}

\subsection{Training Details}
\label{app:train}
For DecSDE and both baseline models, we use the transformer architecture. Both the transformer encoder and decoder have six layers, four attention heads, 512 embedding dimension and 1024 FFN dimension.
All models are trained with a stochastic gradient descent with Adam optimizer, with a learning rate of 5e-4 with a inverse square root scheduler, for a maximum of 50 epochs. Dropout of 0.3, label smoothing of 0.1 are used. These are inherited from fairseq~\cite{fairseq}'s low resource IWSLT'14 German to English (Transformer) example\footnote{https://github.com/pytorch/fairseq/tree/master/examples/translation}. For DecSDE, we have $u=16$ for aze, $u=80$ for bel, $u=0$ for glg and  $u=48$ for slk. 
We select $u$ by manual search based on dev set perplexity. A latent size of 10K is used unless specified otherwise following the original SDE paper. Charater n-gram up to size of 5 are used for aze and glg, and up to 4 for bel and slk. We pick this among 3, 4 an 5 by dev set perplexity. With DecSDE, the models have approximately 60M parameters. In comparison, the baseline LookUp-piece have roughly 52M parameters.
\subsection{Datasets}
TED dataset and preprocessing tools we used are available at https://github.com/neulab/word-embeddings-for-nmt. Futher word segementations are done with SentencePiece running the uni-gram sub-word algorithm.

%\section{Supplemental Material}
%\label{sec:supplemental}

\end{document}